\documentclass{article}
\usepackage{ijcai19}

\usepackage{times}
\usepackage{soul}
\usepackage{url}
\usepackage[hidelinks]{hyperref}
\usepackage[utf8]{inputenc}
\usepackage[small]{caption}
\usepackage{graphicx}
\usepackage{amsmath}
\usepackage{booktabs}
\usepackage{xcolor, soul}
\sethlcolor{yellow}
\usepackage{biblatex} 

\title{Graph Neural Networks in Real-Time Fraud Detection with Lambda Architecture}

\author{
Mingxuan Lu$^1$ \and
Zhichao Han$^2$ \and
Zitao Zhang$^2$ \and 
Yang Zhao$^2$ \and 
Yinan Shan$^2$
\\
\affiliations
$^1$Shanghai Jiaotong University\\
\emails
mingxuan.lu@sjtu.edu.cn

}

\begin{document}
\maketitle

\begin{abstract}
Transaction checkout fraud detection is an essential risk control components for E-commerce marketplaces. In order to leverage graph networks to decrease fraud rate efficiently and guarantee the information flow passed through neighbors only from the past of the checkouts, we first present a novel Directed Dynamic Snapshot (DDS) linkage design for graph construction and a Lambda Neural Networks (LNN) architecture for effective inference with Graph Neural Networks embeddings. Experiments show that our LNN on DDS graph, outperforms baseline models significantly and is computational efficient for real-time fraud detection.
\end{abstract}

\section{Introduction}

Fraudulent transaction is the one of the most serious threats to online security nowadays. This issue is deteriorated by the growing sophistication of business transactions using online payment and payment cards \cite{LAUNDERS2013150,Wang_2021}. Fraudsters apply a range of tactics, including paying with stolen credit cards,
chargeback fraud involving complicit cardholders, selling fake e-gift cards, or creating schemes to create large-scale
layered fraud against multiple merchants\footnote{\url{https://www.verifi.com/in-the-news/need-know-fraud-rings/}}. In this work, we aim to detect risky transaction orders in a real-world e-commerce platform.

Unauthenticated transactions are major buyer risk types to E-commerce marketplaces. It is observed that the entities linking to those orders, such as shipping addresses and device machine ID, are key clues to transaction fraud defection. Hundreds of patterns are summarised as features for models or rules for decision engines. However, the feature engineering of the linkage patterns beyond one-hop on the graph, is currently quite inefficient for human experts.

\textbf{Heterogeneous and Dynamic}. Two vital characteristics of fraudulent transaction orders have been raised \cite{rao2020suspicious} in a similar situation of suspicious massive registration detection. First, transaction orders, as well as the relative entities naturally form a graph with heterogeneous nodes (e.g., IP address, email, accounts) since they tend to share some common risk features, such as the same email or the same IP address. Secondly, the temporal dynamic plays an important role for fraudulent detection because accounts used by fraudsters and by legitimate users usually generate activity events on separate time periods.  \\\\
\textbf{Future information at transaction checkpoints}. Most graph datasets do not consider feature information flowing from the timestamp later than the vertex as a critical issue. For fraud detection, claim, chargeback and suspension history are usually top features for fraud detection models, of which event timestamp matters. Feature patterns from the upcoming events linked by entities may make models "foresee" the risk but this kind of capability is absent in real world checkpoints as future vertices do not appear on the graph yet. \\\\
\textbf{Graph Neighbor Query}. Fraud detection in transaction checkpoint requires low latency response, unlike on-boarding and post-transaction scenarios.  For neighbor query more than 2-hops in graph databases on generally designed linkage relationships, it may last hundreds of milliseconds, which is difficult to meet internal system latency requirements.

In this work, our key contributions are:

\begin{enumerate}
    \item We propose a novel Directed Dynamic Snapshot (DDS) graph design and a Lambda\footnote{Hybrid architecture for learning, batch inference and streaming inference.} Neural Networks (LNN) architecture, which leverages snapshot aggregation, and avoids the model to foresee the information in training.
    \item The LNN on DDS outperforms baseline model in LightGBM significantly, which means graph neighbor and snapshot features are well captured.
    \item LNN together with DDS is suitable for real world low-latency inference as only the last one-hop key-value query is required for graph embedding propagation.
\end{enumerate}

\section{Background} 

In this section, research areas relevant to our work are discussed.

\textbf{GNN}.   Graph neural network (GNN) \cite{Hamilton2017InductiveRL,Kipf2017SemiSupervisedCW,Vaswani2017AttentionIA} has gained incremental popularity in learning from graphs. It has powerful capacity in grasping the graph structure as well as the complex relations among nodes by the means of message passing and agglomeration.

\textbf{TGN on Fraud Detection}.    Dynamic graphs could also be represented as a sequence of time events. Temporal Graph Networks (TGN) \cite{rossi2020temporal} applied memory modules and graph-based operators. The framework of TGN is computationally efficient based on event update. Asynchronous Propagation Attention Network (APAN) \cite{Wang_2021} adopted temporal encoding similar to TGN and decoupled graph computation and inference. However, for TGN, only a small number of neighbors are accessible by graph module due to memory constrains.

\textbf{GNN on Dynamic Graph}.  Learning in temporal dynamic graphs is often set in a scenario of homogeneous graphs. One typical work is DySAT which applies self-attention networks to learn low-dimensional embeddings of nodes in a dynamic homogeneous graph. One notable difference with our setting is that we need distinguish between two types of entities, while DySAT assumes that all entities can be added or removed in the graph.

\textbf{Snapshot GNN on fraud detection}.   
DHGReg \cite{rao2020suspicious} solves suspicious massive registration detection tasks via dynamic heterogeneous graph neural network. DHGReg is composed of two subgraphs, a structural subgraph to reflect the linkages between different types of entities and a temporal subgraph to capture the dynamic perspective of all entities and give different timestamps to different entities as a way to determine whether an entity appears in time \textit{t} or not. With such graph structure, DHGReg manages to grasp the time dimension of heterogeneous graph to detect suspicious massive registered accounts as early as possible. \\\\
In real-world applications, however, issues still remain in the case of DHGReg: (1) the bi-graph structure tends to deplete GPU memory when the graph scale increments; (2) feature information flow from the future to vertices is not constrained; and (3) online neighbor lookup is not effective in deployment.

In this work, we propose LNN on DDS graph to detect suspicious fraud transactions. We adopt merits from the graph structure of DHGReg at the same time coupling both graph computation and online inference in one pipeline. In order to be more compatible with large graphs, we decouple the large graph with partition process before learning. Additionally, we add timestamps to all asset nodes to construct a directed graph only from the effective historical vertices to the target checkout vertices so that the observed feature distribution fits the production scenario better. All assets in the graph is included during graph computation while only one-hop neighbouring entities' embedding are pre-computed after periodical inference and later passed for online inference to decrease inference latency.

\section{Research Question and Methodology}

\subsection{Research Question}

In order to identify transaction risk with graph level information, we would like
to answer the questions below.

\begin{enumerate}
    \item How could we setup linkages between purchase orders
and entities effectively with information from the future of the purchase order creation time excluded?
    \item How could we design the graph neural network architecture which is efficient for online inference?
\end{enumerate}

\subsection{Directed Dynamic Snapshot Graph}

In our experiments, transaction fraud detection is treated as a binary classification problem in inductive setting on a heterogeneous graph.

In a static transaction graph $\mathcal{G}$, a vertex $v \in \mathcal{V}$ has a type $\tau(v) \in \mathcal{A}$, 
where $ \mathcal{A} := \{ order, entity \} $. An edge $e \in \mathcal{E}$ links from an $order$ vertex to an $entity$ vertex.

\begin{table}[]
\centering
\caption{Notations}
\label{tab:notations}
\begin{tabular}{|ll|}
\hline
Notation        & Description                  \\ \hline
$ \mathcal{G} $ & The undirected static graph   \\
$ \mathcal{V} $ & The vertices on static graph \\
$ \mathcal{E} $ & The edges on static graph \\ 
$v$ & An order or entity on the static graph \\
$e$ & Order-entity linkage on the static graph \\
$order$ & Order vertex on static graph \\ 
$entity$ & Entity vertex on static graph \\
$\mathcal{T}$ & The timestamp set \\
$ \mathcal{G_T} $ & The directed dynamic snapshot (DDS) graph   \\
$ \mathcal{G^{E}_T}$ & Effective entity to order graph \\
$ \mathcal{V_T} $ & The vertices on DDS graph \\
$ \mathcal{E_T} $ & The edges on DDS graph \\ 
$order_t$ & Order on snapshot $t$ \\
$order^s_t$ & Shadow Order on snapshot $t$ \\
$entity_t$ & Entity on snapshot $t$ \\
\hline
\end{tabular}
\end{table}

The $order$ nodes with unauthenticated chargeback claims from the customer system are marked as $1$, which are regarded as fraud transactions. The others are marked as $0$, which represent legitimate checkouts. These labels are used for our binary classification problem.

\begin{figure}[htbp]
    \centering
    \includegraphics[width=8cm]{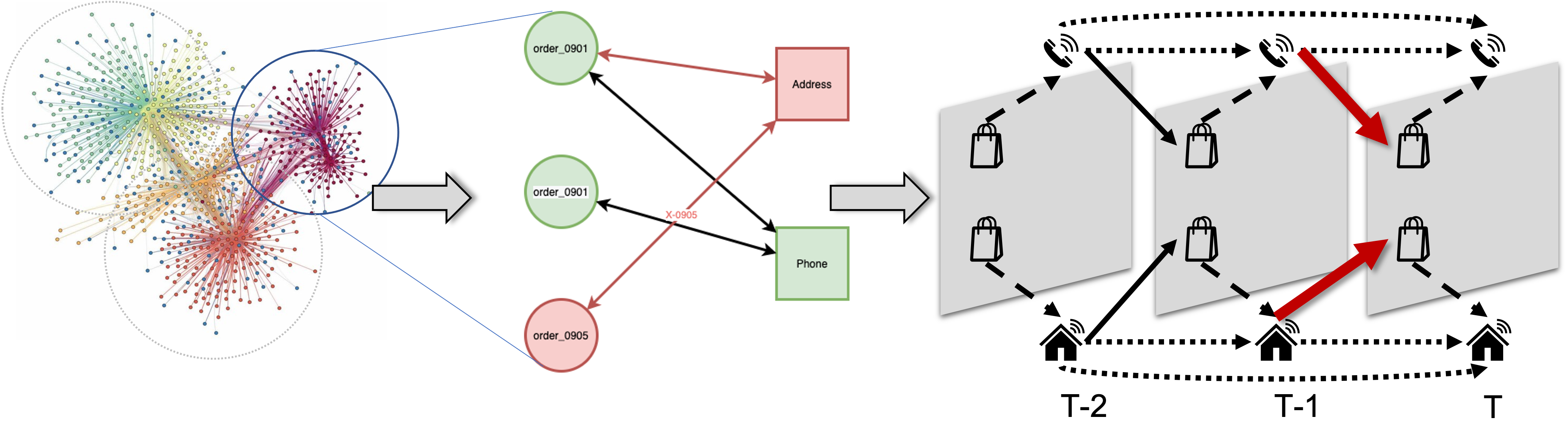}
    \caption{Graph Transform}
    \label{fig:graph-trfm}
\end{figure}

Directed dynamic snapshot graph (DDS) $\mathcal{G_T}$ is transformed from the static transaction graph $\mathcal{G}$ after graph partition as illustrated in Figure \ref{fig:graph-trfm}

A time snapshot $t \in \mathcal{T}$, where $ \mathcal{T} := \{0,1,...,N\} $, could be represented for a period of time duration. e.g. 1 hour and 1 day. In our experiments, the time snapshot represent a day.
A snapshot vertex $v_t \in \mathcal{V_T}$ represents the static vertex which it the snapshot one is transformed from $v$ on snapshot $t$. The edge types for the snapshot vertex linkages are represented in Table \ref{tab:dds-et}.

\begin{table}[]
\centering
\caption{Directed Dynamic Snapshot Graph Edge Types}
\label{tab:dds-et}
\begin{tabular}{|ll|}
\hline
Edge Type  & Description     \\ \hline
$order^s_t \leftrightarrow entity_t$ & Both are in the same $t$ \\ 

$entity_{t-i} \rightarrow entity_t$ &
  Historical entity linkages \\ 
$entity_{t-e} \rightarrow order_t$ &
  Linkages from effective entities  \\ \hline
\end{tabular}
\end{table}

In order to achieve a directed dynamic snapshot graph for GNN to learn from, the graph construction consists of the steps below, illustrated in Fig. \ref{fig:graph-trfm}.

\begin{enumerate}
    \item \textbf{Static Graph Setup} Graph setup based on months of transaction data.
    \item \textbf{Graph Partition} Community detection on transaction graph for learning and inference in parallel.
    \item \textbf{Directed Dynamic Graph Setup} Information flow designed to constrain features extracted from future. 
\end{enumerate}

\subsubsection{Static Graph}

In order to collect neighbor features for transaction fraud risk evaluation, multiple entities directly used in checkout sessions are adopted as $order$ neighbors. These entities including shipping address, E-mail, IP address, device ID, contact phone, payment token and user account, are represented as $entity$ nodes on $\mathcal{G}$.

Each $order$ vertex represents a checkout transaction along with an unique order ID, 
linked with multiple $entity$ vertices 
such as shipping addresses, E-mails, contact phones that buyers need to confirm in checkout pages.
Most of $entity$ vertices are linked to multiple $order$ vertices as well.

\subsubsection{Graph Partition}

The static graph is generated from months of transactions so as to get fewer stand-alone $entity$ vertices and make the information passed through neighbors effective. However, on the other hand, 
the linkages coming from a long time period creates a huge graph with billions of nodes with dozens of extremely large connected components. It makes deep learning and model inference in parallel difficult from the unprocessed graph. 

Power Iteration Clustering (PIC) \cite{DBLP} is utilized to partition the graph, which ensures the graph connectivity and reduces the graph sparsity effectively on extreme large graphs. It is a handy approach as in COMPANY, thousands of ETL jobs run on Apache Spark, where PIC is a built-in algorithm.

In order to keep the community size close to the business understanding for a gang of fraudsters, which is around 1000, the cluster results from PIC are further processed with METIS \cite{karypis1998fast}. It makes the graph learning on the mini communities afterwards in  ClusterGCN \cite{Chiang_2019} flavor.

\subsubsection{Directed Dynamic Snapshot Graph}

After graph partition, the static graph is decoupled into thousands of small communities. 
The small static communities are not chosen to apply GNN as the inference score on $order$ on timestamp $t$ may obtain embeddings coming from the future snapshot $t+1$. The embeddings from future is a critical issue to fraud detection as it makes the detector foresee some kind of facts from timestamp $t+1$ summarised in the embeddings which should not be observable at timestamp $t$.

In order to constrain the embeddings seen for $order$ at timestamp $t$ obtains only the information from the past, DDS graph $\mathcal{G_T}$ is utilized. The workflow is described as below:

\begin{enumerate}
    \item Construct $order_t$ from $order$ as effective order, along with a label for learning and evaluation.
    \item Clone shadow $order^s_t$ from $order_t$, which is not linked with any label and get engaged in interaction with entities in the same snapshot.
    \item Create $entity_t$ from $entity$, which represents the entity in snapshot.
    \item Link $order^s_t$ and $entity_t$ to make information flow between them. For vertex $order_{t+1}$, if $order^s_t$ and $order_{t+1}$ are both linked with the same entity, the shadow $order^s_t$ will act as the past order to the $entity_{t+1}$ and $order_{t+1}$ will be latest one, which is the order to be evaluated. The shadow $order^s_t$ is introduced because the information do not flow between $order_t$ and $entity_t$.
    \item Create edges from $entity_{t-i}$ to $entity_t$, where $0 \leq t-i \leq t$. These edges represents information flow from the past entity to the current entity and the self-loop on the current entity. $entity_t$ may be connected with a bunch of $entity_{t-i}$ as long as they are linked to any $order_{t-i}$.
    \item Create edges from $entity_{t-e}$ to $order_t$, where $0 \leq t-e < t$. $order_t$ has only one edge to the identical effective entity for each entity type such as phone and email. These edges are the final 1-hop edges from the $order_t$ which is to be evaluated or learnt along with labels. These edges are the only edges required for production online inference, which simplifies the neighbor lookup in graph databases into embedding lookup in key-value databases.
\end{enumerate}

\begin{figure}[htbp]
    \centering
    \includegraphics[width=8cm]{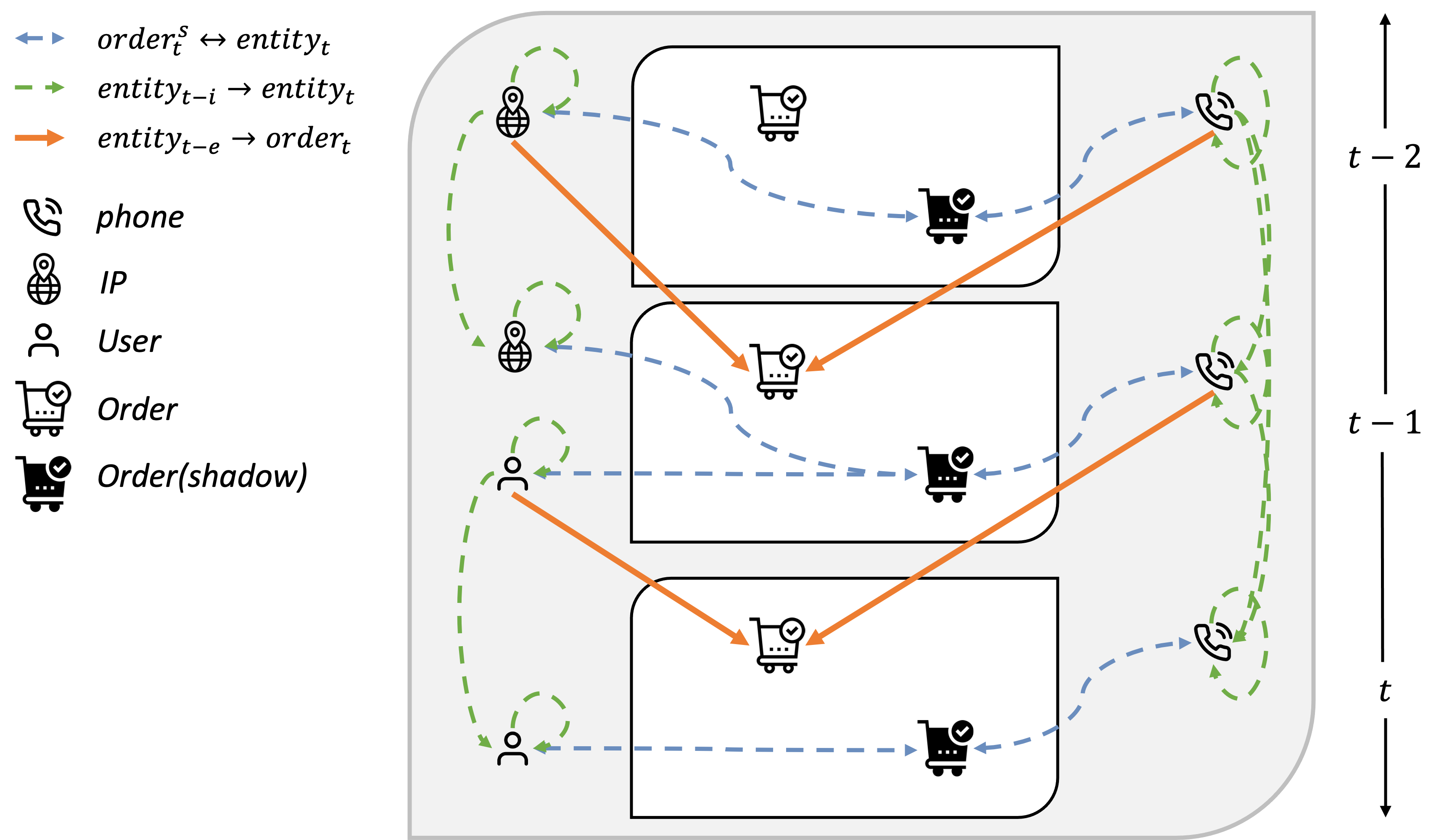}
    \caption{Directed Dynamic Snapshot Graph}
    \label{fig:graph-dds}
\end{figure}

With DDS graph, displayed in Fig. \ref{fig:graph-dds}, information coming from the past for $order_t$ is guaranteed, which answers research question 1.

\subsection{Network Architecture}

In order to leverage the last edge from DDS, a two-stage, or Lambda neural network (LNN) architecture is proposed so as to reuse embeddings from $entity_{t-e}$ without querying $k$-hop neighbors. Similar to two-tower
deep learning model architecture \cite{covington2016deep}, in which user-profile and item embeddings are usually pre-computed, the embeddings of entities, such as E-mail and contact phones, are inferred before-hand and fetched from key-value stores for linked purchase orders before checkout approvals. The LNN architecture, togheter with DDS graph, answers the research question 2, which is the solution proposed for online inference with graph structure feature aggregated.

\begin{figure}[htbp]
    \centering
    \includegraphics[width=8cm]{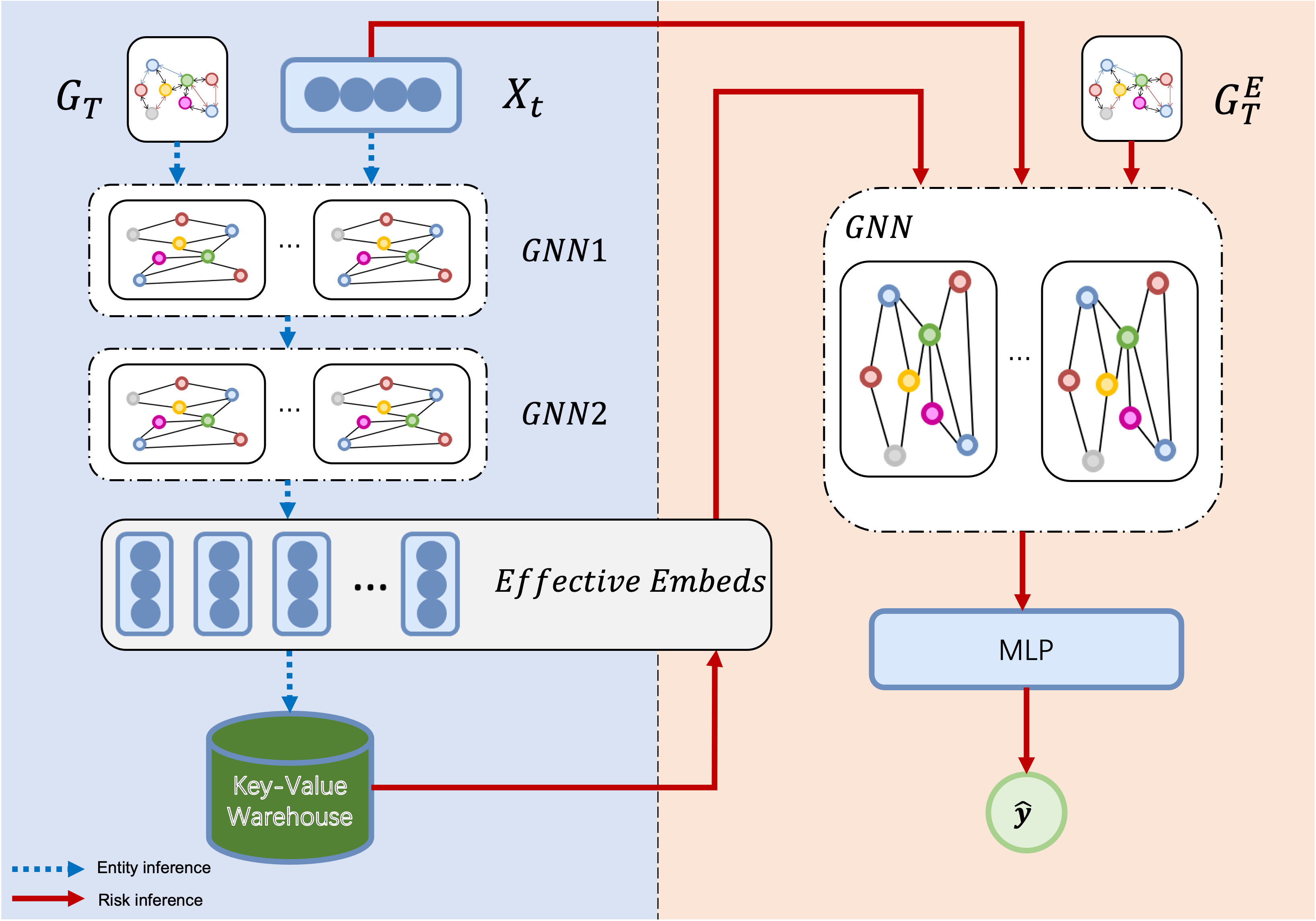}
    \caption{Lambda Network Structure}
    \label{fig:LNN}
\end{figure}

As illustrated in Figure \ref{fig:LNN}, LNN consists of a block of GNN layers block and a Multilayer Perceptron (MLP), similar to DeepGCNs \cite{li2019deepgcns}. The two stages are splitted from the whole model at the vertex last effective vertex $entity_{t-e}$. The first stage is the head block of GNN layers except the last layer, of which the final layer embedding output represents the latest effective $entity_{t-e}$. For the second stage, namely, the one-layer GNN followed by the MLP, takes the embeddings of $entity_{t-e}$ and raw features of $order_t$.

\subsubsection{Training phase}

In training phase, both stages are used for end-to-end learning. The information flowing from historical shadow $order^{t-i}$ to latest effective $entity^{t-e}$, then from $entity^{t-e}$ to $order_t$ with labels, without the information later than $order_t$.

\subsubsection{Entity Embedding Inference}

In production environment, $entity_{t-e}$ vertex embeddings are periodically refreshed from DDS graphs with a fixed number of snapshots by the first stage of LNN. The values could be stored in distributed key-value stores for multiple downstream purposes. In our paper, the embeddings focus on transaction fraud risk detection.

\subsubsection{Transaction Risk Inference}

Before checkout approval, the second stage of LNN evaluate the score from the entity embeddings and raw purchase order features, which is equivalent to the risk score inferred from the unprocessed historical connected transaction features from scratch, given the linkages from DDS graph.


\section{Evaluation}

\subsection{Dataset}

The static graph is constructed from months of transactions and only entities used in checkout sessions are adopted.

\subsection{Experimental Setup}

The experiments are processed in training phase. The training data comes from the checkouts in the first 80\% time snapshots. The middle 10\% are used as validation set for early stopping. The final 10\% of the snapshot orders are used for testing. 

Graph partition is processed on the whole static graph without stages for training, validation or testing. The number of expected partition number is set based on expected partition size. The expected partition size for PIC \cite{DBLP} is 1 million. The expected partition size got from METIS \cite{karypis1998fast} is set to 1024. It would be interesting to further explore how could the partition size impact our model performance. For the experiments in this paper, the numbers are set based on business advice.

Directed graph edges in $\mathcal{G_T}$ and checkout features on $order_t$ are used for GNN blocks. For $entity_t$ vertices, the initial features are set to zero, which makes the LNN model performance comparison to models without $entity$ information fair.

Gradient Boosting Decision Tree models \cite{dorogush2018catboost,ke2017lightgbm} are still first choices models in fraud detection domains due to its outstanding performance on tabular datasets, as which our raw checkout features are represented. For MLP and LNN training, we use the encoded features from an existing LightGBM (LGB) \cite{ke2017lightgbm} trained for risky transaction detection.

\subsection{Results}

We report Average Precision (AP) and Area Under the Receiver Operating Characteristic Curve (ROC AUC) for the predicted scores in Table \ref{tab:exp-sim}. As is shown, LNN with feature aggregated through graph linkages, outperforms MLP significantly both in ROC AUC and AP. Compared with the LGB, which is still the state-of-art model for tabular feature set, LNN achieves 49.22\%, which is 9\% higher than the AP obtained from LGB.

\begin{table}[]
\centering
\caption{Experiment Results}
\label{tab:exp-sim}
\begin{tabular}{|lll|}
\hline
Model & ROC AUC & Average Precision \\ \hline
MLP   & 0.9217±0.0014   & 0.3912±0.0029             \\
LGB   & 0.9317±0.0005   & 0.4081±0.0096             \\ \hline
LNN (GAT) & 0.9381±0.0012   & 0.4755±0.0100         \\ 
LNN (GCN) & 0.9431±0.0008 & 0.4922±0.0024 \\ \hline
\end{tabular}
\end{table}

\section{Conclusions}

We present an approach LNN on DDS graph, to leverage both graph structure and time snapshot features for fraud transaction detection. The LNN outperforms LGB significantly, which is promising to help risk data scientists get rid of tedious graph feature engineering. Information from the future of the transaction is well constrained helps to reduce the gap between experiment and production feature distributions. LNN, together with DDS graph design, requires only 1-hop embedding value query, makes low-latency inference feasible.

LNN is a generic GNN blocks, as long as the input graph follows the DDS graph design principles. With more representative GNN layers, it is promising the achieve better performance.

\printbibliography

@article{Kipf2017SemiSupervisedCW,
  title={Semi-Supervised Classification with Graph Convolutional Networks},
  author={Thomas Kipf and M. Welling},
  journal={ArXiv},
  year={2017},
  volume={abs/1609.02907}
}

@article{Vaswani2017AttentionIA,
  title={Attention is All you Need},
  author={Ashish Vaswani and Noam M. Shazeer and Niki Parmar and Jakob Uszkoreit and Llion Jones and Aidan N. Gomez and Lukasz Kaiser and Illia Polosukhin},
  journal={ArXiv},
  year={2017},
  volume={abs/1706.03762}
}

@inproceedings{Hamilton2017InductiveRL,
  title={Inductive Representation Learning on Large Graphs},
  author={William L. Hamilton and Z. Ying and J. Leskovec},
  booktitle={NIPS},
  year={2017}
}

@misc{rao2020suspicious,
      title={Suspicious Massive Registration Detection via Dynamic Heterogeneous Graph Neural Networks}, 
      author={Susie Xi Rao and Shuai Zhang and Zhichao Han and Zitao Zhang and Wei Min and Mo Cheng and Yinan Shan and Yang Zhao and Ce Zhang},
      year={2020},
      eprint={2012.10831},
      archivePrefix={arXiv},
      primaryClass={cs.LG}
}

@article{rossi2020temporal,
  title={Temporal graph networks for deep learning on dynamic graphs},
  author={Rossi, Emanuele and Chamberlain, Ben and Frasca, Fabrizio and Eynard, Davide and Monti, Federico and Bronstein, Michael},
  journal={arXiv preprint arXiv:2006.10637},
  year={2020}
}

@article{Wang_2021,
   title={APAN: Asynchronous Propagation Attention Network for Real-time Temporal Graph Embedding},
   ISBN={9781450383431},
   url={http://dx.doi.org/10.1145/3448016.3457564},
   DOI={10.1145/3448016.3457564},
   journal={Proceedings of the 2021 International Conference on Management of Data},
   publisher={ACM},
   author={Wang, Xuhong and Lyu, Ding and Li, Mengjian and Xia, Yang and Yang, Qi and Wang, Xinwen and Wang, Xinguang and Cui, Ping and Yang, Yupu and Sun, Bowen and et al.},
   year={2021},
   month={Jun}
}

@inproceedings{DBLP,
  author={Frank Lin and William W. Cohen},
  title={Power Iteration Clustering},
  year={2010},
  cdate={1262304000000},
  pages={655-662},
  url={https://icml.cc/Conferences/2010/papers/387.pdf},
  booktitle={ICML},
  crossref={conf/icml/2010}
}

@incollection{LAUNDERS2013150,
title = {Chapter 13 - A Semantic Approach to Security Policy Reasoning},
editor = {Babak Akhgar and Simeon Yates},
booktitle = {Strategic Intelligence Management},
publisher = {Butterworth-Heinemann},
pages = {150-166},
year = {2013},
isbn = {978-0-12-407191-9},
doi = {https://doi.org/10.1016/B978-0-12-407191-9.00013-2},
url = {https://www.sciencedirect.com/science/article/pii/B9780124071919000132},
author = {Ivan Launders and Simon Polovina}
}

@article{karypis1998fast,
  title={A fast and high quality multilevel scheme for partitioning irregular graphs},
  author={Karypis, George and Kumar, Vipin},
  journal={SIAM Journal on scientific Computing},
  volume={20},
  number={1},
  pages={359--392},
  year={1998},
  publisher={SIAM}
}

@article{Chiang_2019,
   title={Cluster-GCN},
   ISBN={9781450362016},
   url={http://dx.doi.org/10.1145/3292500.3330925},
   DOI={10.1145/3292500.3330925},
   journal={Proceedings of the 25th ACM SIGKDD International Conference on Knowledge Discovery \& Data Mining},
   publisher={ACM},
   author={Chiang, Wei-Lin and Liu, Xuanqing and Si, Si and Li, Yang and Bengio, Samy and Hsieh, Cho-Jui},
   year={2019},
   month={Jul}
}

@inproceedings{covington2016deep,
  title={Deep neural networks for youtube recommendations},
  author={Covington, Paul and Adams, Jay and Sargin, Emre},
  booktitle={Proceedings of the 10th ACM conference on recommender systems},
  pages={191--198},
  year={2016}
}

@misc{li2019deepgcns,
      title={DeepGCNs: Can GCNs Go as Deep as CNNs?}, 
      author={Guohao Li and Matthias Müller and Ali Thabet and Bernard Ghanem},
      year={2019},
      eprint={1904.03751},
      archivePrefix={arXiv},
      primaryClass={cs.CV}
}

@article{ke2017lightgbm,
  title={Lightgbm: A highly efficient gradient boosting decision tree},
  author={Ke, Guolin and Meng, Qi and Finley, Thomas and Wang, Taifeng and Chen, Wei and Ma, Weidong and Ye, Qiwei and Liu, Tie-Yan},
  journal={Advances in neural information processing systems},
  volume={30},
  pages={3146--3154},
  year={2017}
}

@article{dorogush2018catboost,
  title={CatBoost: gradient boosting with categorical features support},
  author={Dorogush, Anna Veronika and Ershov, Vasily and Gulin, Andrey},
  journal={arXiv preprint arXiv:1810.11363},
  year={2018}
}
\end{document}